%% file: main.tex
\documentclass[a4paper]{article}

\usepackage{INTERSPEECH2020}
\usepackage{comment}
\usepackage{xcolor,enumitem}

\title{Analysis of Disfluency in Children's Speech}
\name{Trang Tran$^1$*\thanks{\ \ *Equal Contribution.}, 
Morgan Tinkler$^2$*, 
Gary Yeung$^2$, 
Abeer Alwan$^2$, 
Mari Ostendorf$^1$}
\address{
  $^1$Dept. of Electrical and Computer Engineering, University of Washington, USA \\
  $^2$Dept. of Electrical and Computer Engineering, University of California Los Angeles, USA}
  
\email{\{ttmt001, ostendor\}@uw.edu, \{mckeatink, garyyeung\}@g.ucla.edu, alwan@ee.ucla.edu}

\begin{document}

\maketitle
\begin{abstract}
Disfluencies are prevalent in spontaneous speech, as shown in many studies of adult speech. Less is understood about children's speech, especially in pre-school children who are still developing their language skills. We present a novel dataset with annotated disfluencies of spontaneous explanations from 26 children (ages 5--8), interviewed twice over a year-long period. Our preliminary analysis reveals significant differences between children's speech in our corpus and adult spontaneous speech from two corpora (Switchboard and CallHome). Children have higher disfluency and filler rates, tend to use nasal filled pauses more frequently, and on average exhibit longer reparandums than repairs, in contrast to adult speakers.
Despite the differences, an automatic disfluency detection system trained on adult (Switchboard) speech transcripts performs reasonably well on children’s speech, achieving an F1 score that is 10\% higher than the score on an adult out-of-domain dataset (CallHome).
\end{abstract}
\noindent\textbf{Index Terms}: children's speech, disfluency, acoustic analysis, fundamental frequency

\input{intro}

\input{corpus}

\input{analysis}

\input{compare-adult}

\input{conclusion}

\section{Acknowledgements}
This work was supported in part by National Science Foundation (NSF) Grant \#1734380.

\bibliographystyle{IEEEtran}
\bibliography{mybib}


\end{document}

%% file: intro.tex
\section{Introduction}

Disfluencies, including filled pauses, repetitions and self-corrections, are common in spontaneous speech. As speech-based communication with devices and virtual agents becomes more natural, it will be increasingly important for conversational agents to detect and use disfluencies in understanding users.
While there have been many studies of disfluencies in adult spontaneous speech, including extensive work on automatic disfluency detection, most studies of children speech have focused on clinical applications.  
Understanding child speech disfluency is important in evaluation of language development. Analyzing speech characteristics in children can help distinguish disfluencies that are natural in typical development vs.\ signs of autism spectrum disorder \cite{MacFarlane2017}, attention-deficit/hyperactivity disorder \cite{Lee2018}, or language disorders \cite{Bergstrom2017}, most commonly stuttering \cite{Yaruss1999,Tumanova2014,Hollister2017}. 

In addition, it can be useful to detect disfluencies in non-clinical contexts. 
For example, automatic detection of disfluencies in read speech is useful for assessing a child's reading ability \cite{Proenca2015}.
Social companion robots show promise as both assessment tools and educational partners for young children \cite{Sadeghian2015, Spaulding2018, Yeung2019}. In this context, or for children talking to virtual agents more generally, disfluency detection is needed to facilitate automatic speech understanding and assessing child engagement.


Few corpora of spontaneous children’s speech are available, and even fewer exist with annotated disfluencies. There is some data for read speech, e.g.\ \cite{Proenca2015,chorec-corpus}, and a corpus of child-computer interaction \cite{Yildirim09}. The work in this paper contributes a novel dataset of transcripts of human-directed spontaneous speech from children with disfluency annotations,\footnote{For privacy reasons, only annotated transcripts are made available: {\tt www.seas.ucla.edu/spapl/shareware.html}}
together with distributional analyses and automatic detection results.





%% file: corpus.tex
\section{Corpus and Annotation Description}
\label{sec:corpus}
The dataset developed for this study is based on a set of interviews between an adult and a child, using a protocol described in \cite{Bailey2014,Bailey2019}.
The data collected is part of a larger human-robot interaction (HRI) study evaluating the effectiveness of social robots in classroom settings \cite{Yeung2019, Yeung2019b}.
The robotic medium is JIBO, a social robot originally developed to be a home personal assistant \cite{Jibo2017}. JIBO was designed to act as a learning companion, serving as the child's peer with a friendly child-like voice. 
We chose a subset of interactions with a human teacher in order to compare results to adult conversational data, and because our goal is to support more human-oriented interactions.

\subsection{Data collection}
\label{ssec:data-collection}


A microphone was placed between the teacher and the child, at a 45$^{\circ}$ angle approximately 30-50 cm away from both participants. 
26 children (15 female and 11 male) were each interviewed twice, approximately one year apart, ages 4.8 to 7 in the first interview. Overall the dataset consisted of 7 hours of recorded and transcribed interviews, reduced to approximately 1.26 hours of child speech. Each interview consisted of a series of questions designed to elicit spontaneous explanations from the children through a narrative task. 

During the first interview, the children were prompted regarding two tasks: brushing their teeth (`teeth 1') and mixing paint into colors (`colors'). They were asked: 1) how they accomplished this task, 2) why they should perform this task, 3) how to explain the task to a friend, and 4) why that friend should perform the given task the way they do. 
During the second interview, the children were prompted with three tasks. First, they were presented a series of four photos of different animals and asked to identify which animal was the odd one out and explain why (`animals'). Second, the teeth-brushing task was repeated (`teeth 2'). Third, the children were presented with an unknown number of cubes that could be either attached to one another or split apart and then asked to identify how many cubes they had been given (`blocks'). The same series of questions (1--4) were then asked about this new counting task. 


   
\subsection{Annotation Process}
\label{ssec:annotation-process}
The annotation framework builds on standards developed for adult speech used on the Switchboard corpus \cite{Godfrey1993}, including disfluencies, indication of fillers, and segmentation boundaries. We incorporate minor modifications and add markers for child hesitations and partner backchannels. As in other spontaneous speech corpora, some segments of speech are difficult to understand and are labeled as `[inaudible]'. The conventions were chosen by three annotators, after multiple sessions of listening to and annotating seven audio files. 
Figure~\ref{fig:example} provides an example child dialog associated with a protocol used for all children in both sessions, illustrating most of the annotated phenomena.

Disfluencies included repetitions, restarts, and self-repairs, which reflect production/planning issues.
The disfluency notation chosen builds on the annotation standard originally outlined in \cite{shriberg94}. Specifically, a disfluency consists of a reparandum followed by an interruption point '+', an optional interregnum '\{xx\}', and then the repair, if any. A few simple examples of adult disfluencies are given below:
\begin{verbatim}
  [was it + {I mean} did you] put...
  [I just + I] enjoy working...
  [By + ] it was attached to...
\end{verbatim}
We use a variant that omits the nested bracketing structure for repetition disfluencies proposed in \cite{Hahn2013}, e.g., using ``[he + he run + he run]'' as opposed to ``[he + [he run + he run]].'' Disfluencies sometimes involve word fragments, which are transcribed with a final hyphen, as in ``[b- + b- + but]'' or ``[he w- + he put].''

Fillers are words that are used to hold the floor while one is thinking and can be removed without affecting the meaning of a sentence. Filler words or phrases do not include discourse markers such as `so' or `well' or agreement backchannels such as `yeah' or `right.' In Switchboard annotations, fillers mainly include the filled pauses `uh' and `um.' For the child disfluency corpus, we included words such as `like' as fillers. This may reflect a difference in conventions, or simply a difference in language use, since the Switchboard data was collected roughly 30 years ago. Fillers are indicated with `\{F xx\}' notation. In Switchboard, fillers are typically associated with the interregnum in a disfluency. In the child data, if the pause occurs after the filler, we associate the filler with the reparandum.

Segmentation boundaries include turn boundaries (indicated by `//') and sentence-like unit (SU) boundaries (marked with `/'). Turn boundaries separate full speaker turns. SU boundaries indicate semantically coherent units within a speaker’s turn, allowing for the fact that spontaneous speech does not always result in grammatical sentences. Each SU conveys a complete meaning or speech act, which might be a simple noun or verb phrase in answer to a question. In spontaneous speech, clauses that start with a conjunction are often considered a single SU.

Another phenomenon that was frequent in the child data was hesitation indicated with an unfilled pause and/or duration lengthening that was not perceived as fluent. These word boundaries, indicated with `\{H\},' are not used for pauses or prolongations that occur at SU boundaries, interruption points, or words that are lengthened for emphasis. 
These annotations are included in the corpus, but excluded from analysis as the inter-annotator agreement was not high 
(see \ref{ssec:annotator-agreement}). 

The instructor speech was not transcribed in our dataset, since it primarily followed a prescribed script. However, we decided to annotate backchannels, referred to as partner backchannels (denoted `\{PBC\}'), since these tended to occur at points of hesitation and SU boundaries associated with child uncertainty. They represent encouragement for the child to continue. 


\begin{figure}
\begin{tabular}{ll}
A:& {\sl Tell me how you clean your teeth.} \\
{\color{blue} C}:& {\color{blue}by brushing \{H\} your tooth \{PBC\} //} \\
A:& {\sl Okay. Anything else you can tell me about
   how you} \\
   & {\sl clean your teeth?}\\
{\color{blue} C}:& {\color{blue}[you + you] get a brush [and then s- + and 
   then put] it}\\
   & {\color{blue}and [some + some] [like + like] just squeeze it / and} \\
   & {\color{blue}[then + then] you put 
   a little bit of water on it \{PBC\} /} \\
   & {\color{blue}and then you brush your teeth / 
and then you spit it out /} \\
   & {\color{blue}and then you get more water like this / and then you} \\
   & {\color{blue}drink it / and then you spit it out again //} \\
A:& {\sl Okay. Now tell me why you clean your teeth.} \\
{\color{blue}C}:& {\color{blue}[because i +] it’s very important / [so i + so i] can} \\
 & {\color{blue}eat bubblegum [all + all] the time //}  \\
A:& {\sl Okay, anything else you can tell me about why you} \\
 & {\sl clean your teeth?} \\
{\color{blue}C}:& {\color{blue}\{F mh\} [because + because] so you can't have germs}\\
 & {\color{blue}anymore / so you can eat bubblegums} //    
\end{tabular}
\caption{Example dialog. A=adult; C=child}
\label{fig:example}
\end{figure}

Annotations were made in all lower case and without punctuation. 
Some speech patterns were not captured by the annotation, such as tongue clicking, nasal speech, exasperated replies, whispered replies, 
and the replacement of fricatives with stops.

\subsection{Inter-Annotator Agreement}
\label{ssec:annotator-agreement}
Inter-annotator agreement was measured between two annotators over 15 files (3,700 tokens). For boundary agreement, annotations were first compared for 5 categories: None, \{H\}, +, /, and //. Cohen's kappa for these 5 categories was 0.71. The agreement for unfilled pauses 
was particularly low: the two annotators both identified the unfilled pause in only 32 tokens, but disagreed on the presence/absence of \{H\} for 112 tokens. Therefore, unfilled pauses were excluded from later analyses. With the remaining 4 boundary categories, inter-annotator agreement was reasonably high with Cohen's kappa of 0.77.
For disfluency annotations, agreement was compared based on binary labels of whether or not a token was in a reparandum. Cohen's kappa for disfluency annotation was very high at 0.82, indicating a high reliability for identification of disfluencies.

%% file: analysis.tex
\section{Child Speech Data Analysis}

\subsection{Transcription Analysis}
\label{ssec:analysis-stats}
Table \ref{tab:gender-stats} presents disfluency statistics in our dataset. Statistical significance for rate differences between groups was assessed using the Poisson e-test \cite{poisson-etest}; a t-test was used for length differences. 
Overall disfluency and filler rate is relatively high at 15.2\%, compared to results reported for adult speech (see Section \ref{ssec:adult-stats}).
It is also high relative to the average of 7.4\% reported in \cite{Yildirim09} for a collection of speech from 10 children ages 4-6 recorded in dialogs with a computer agent. This is consistent with the observation that adult disfluency rates are higher in human-human conversations compared to human-computer conversations.

Comparing between genders, notable differences are: female children tend to be less disfluent (8.5\% vs.\ 12.1\%) and use fewer fragmented words (1.2\% vs.\ 2.5\%), but use fillers more frequently (5.4\% vs.\ 4.5\%) than male children. Higher disfluency rates in male children is consistent with findings in \cite{McLaughlin1989,Tumanova2014} but is in contrast to a study on Switchboard adult speech \cite{Shriberg96}, where it was observed that men had higher filler rates, hypothesized as a strategy for floor-holding. 

\begin{table}[hbtp]
\centering
\caption{Disfluency statistics in the child speech corpus: overall and comparing between genders. {\bf Bold} denotes statistically significant difference between genders at $p<0.05$.}
\label{tab:gender-stats}
\begin{tabular}{lccc}
\toprule
                & overall  & female (2x15) & male (2x11) \\
\midrule
\# tokens          & 13,568      & 7436      & 6132      \\
\# turns           & 2,119       & 1201      & 918       \\
avg. SU length     & 6.4.        & {\bf 6.2}      & {\bf 6.7}       \\
disf. rate         & 10.1\%      & {\bf 8.5\%}    & {\bf 12.1\%}    \\
filler rate        & 5.0\%       & {\bf 5.4\%}     & {\bf 4.5\%}    \\
\% filler in disf. & 12.1\%      & 13.3\%    & 10.2\%    \\
`uh' rate          & 0.5\%       & 0.6\%     & 0.5\%     \\
\% `uh' in disf.   & 16.2\%      & 13.3\%    & 20.7\%    \\
`um' rate          & 2.3\%       & {\bf 2.6\%}     & {\bf 1.9\%}     \\
\% `um' in disf.   & 14.4\%      & 14.4\%    & 14.4\%    \\
frag. rate         & 1.8\%       & {\bf 1.2\%}     & {\bf 2.5\%}     \\
\bottomrule
\end{tabular}
\end{table}

\begin{figure}[hbpt]
  \centering
  \includegraphics[width=\linewidth]{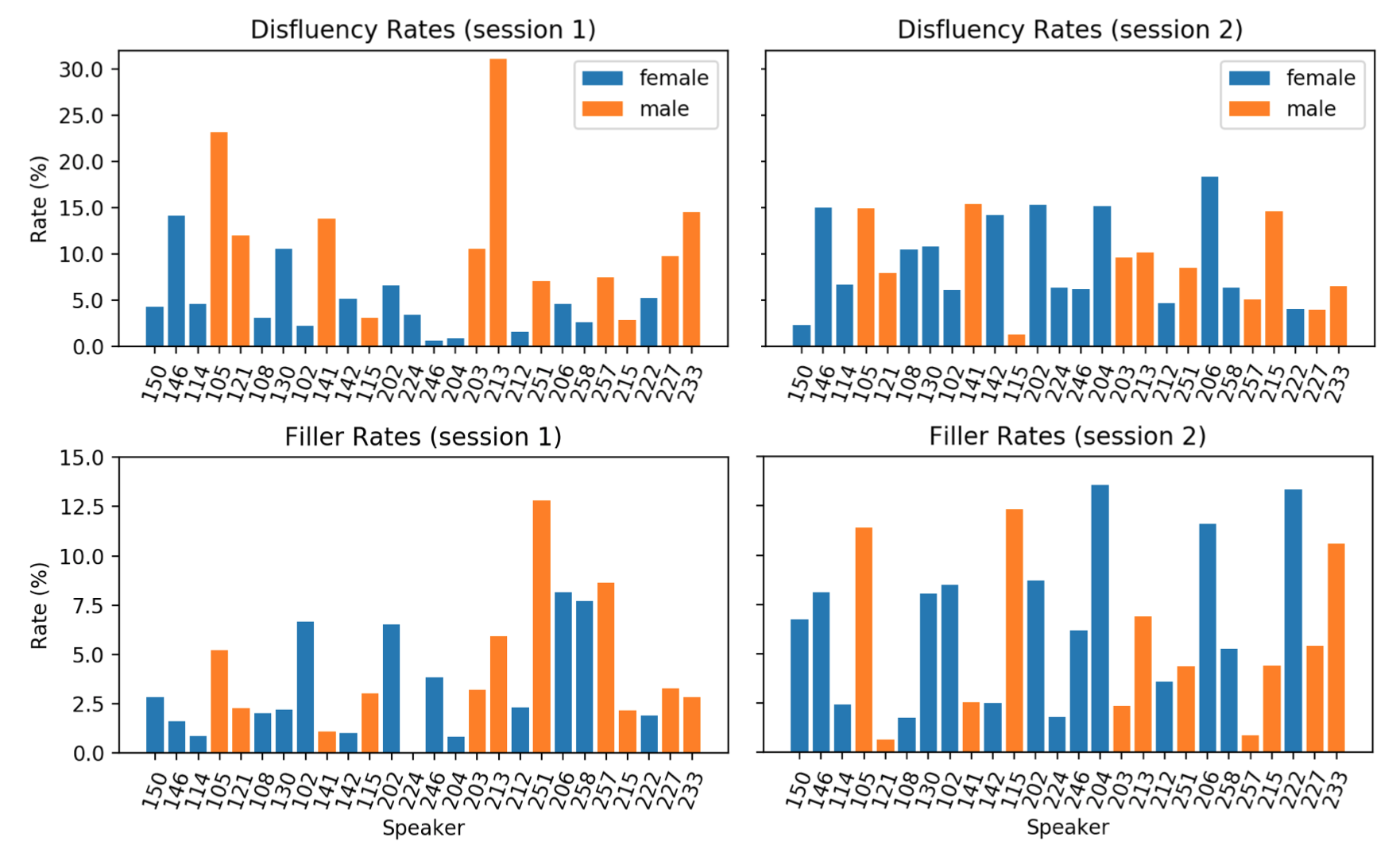}
  \caption{Disfluency and filler rates by speaker, for both sessions. Speakers are sorted by age at the time of the first session (from low to high, left to right).}
  \label{fig:dff-filler-rates}
\end{figure}

Figure \ref{fig:dff-filler-rates} shows disfluency and filler rates by child for the two sessions, ordered by first session age. Averaging over the two sessions, the child disfluency rate ranges from 2.3\% to 20.6\% ($\mu = 9.0, \sigma = 4.9$), and the filler rate ranges from 0.9\% to 10.8\% ($\mu = 5.2, \sigma = 2.7$). For comparison, the rates for disfluencies and fillers together for the 10 children in \cite{Yildirim09} range from 3.3\% to 13.6\%.


Comparing between two interview sessions, overall there was no significant difference in disfluency rate (9.7\% vs\ 10.4\%) but the higher filler rate in the later session (6.0\% vs. 3.6\%) was statistically significant ($p<0.05$). Comparing across tasks (Table \ref{tab:compare-tasks}), the task that stood out was the odd-one-out animal task, where children responded with significantly shorter segments and higher filler rate than in other tasks. This result might be due to the more challenging nature of the `animals' task. 

The teeth-brushing task was common between the two sessions. There was a surprising difference in both disfluency and filler rates, with the second session again having higher rates. 
The increase is observed for 17 of the 26 speakers and is statistically significant in aggregate ($p<0.05$). 
The reason is unclear.
One hypothesis is that the teeth-brushing task in the first interview was conducted first, while in the second interview it was after the `animals' task, priming the children at a higher cognitive load.

\begin{table}[hbpt]
\centering
\caption{Disfluency statistics across different tasks.
{\bf Bold} denotes statistically significant difference between the group and the rest of the groups at $p<0.05$.}
\label{tab:compare-tasks}
\begin{tabular}{lccccc}
\toprule
               & teeth 1 & teeth 2 & colors & animals & blocks \\
\midrule
\# tokens      & 2617    & 3496    & 2870   & 1179    & 3406   \\
\# turns       & 416     & 532     & 453    & 206     & 512    \\
SU len.      & 6.3     & 6.6     & 6.3    & {\bf 5.7}     & 6.7    \\
disf. rate     & {\bf 8.9\%}   & 11.3  & 10.4\% & {\bf 8.2\%}   & 10.2\% \\
filler rate    & {\bf 3.2\%}   & {\bf 5.7\%}   & {\bf 3.9\%}  & {\bf 7.5\%}   & {\bf 5.8\%}  \\
frag. rate     & 2.0\%   & 1.8\%   & 2.2\%  &  1.2\%   & 1.6\%  \\
\bottomrule
\end{tabular}
\end{table}

\subsection{Acoustic Analysis}
\label{ssec:analysis-acoustics}
Automatic word-level forced alignment was performed using a time delay neural network hidden Markov model automatic speech recognition system with approximately 6000 senones \cite{Peddinti2015}.
The system was implemented using Kaldi \cite{Povey2011} and trained on approximately 90 hours of child speech in various classroom settings from the TBALL Children's Speech Corpus \cite{Kazemzadeh2005}.
Due to the difficulty of child speech forced alignment, we compared human-annotated time alignments of 3 conversations of the teeth-brushing task against the automatic forced alignment system.
For these conversations, the differences in the marked word boundaries between the human and automatically generated alignments differed by an overall average of 300ms, with the largest errors contributed by words at turn boundaries and interfering loud background noise. Excluding these few problematic tokens, the average difference between human and automatic alignments was 95ms.

To extract the fundamental frequencies ($f_0$) of child speech, several pitch detection algorithms were evaluated, including BaNa pitch estimation \cite{Ba2012}, multi-band summary correlogram (MBSC)-based pitch estimation \cite{Tan2013}, and sawtooth waveform inspired pitch estimation (SWIPE) \cite{Camacho2008}. After inspection, we observed no significant difference between the three algorithms in both pitch estimation and voiced frame detection across 3 conversations.
For this study, we used MBSC-based $f_0$ estimation.
We assessed the MBSC-based $f_0$ by inspecting 150 frames of voiced speech across the 3 conversations, and the
relative difference between the human annotated and automatically estimated $f_0$ was less than 6.7\%.


\begin{table}[hbpt]
\centering
\caption{Average mean and standard deviation of $f_0$ (Hz) for each token category, separated by age.}
\label{tab:age_c_averages}
\begin{tabular}{lllllll}
\toprule
Word Category   & [4.8-6) &  [6-7) &  [7-8)  \\
\midrule
filler       & 241$\pm$15  & 238$\pm$18 & 223$\pm$13     \\
filler@boundary & 249$\pm$14        & 249$\pm$17 & 218$\pm$13      \\
fluent        & 250$\pm$9        & 245$\pm$10      & 227$\pm$9      \\
boundary        & 249$\pm$21        & 246$\pm$21      & 228$\pm$18      \\
interruption point         & 251$\pm$16        & 243$\pm$10      & 225$\pm$11      \\
within disf.  & 251$\pm$10        & 249$\pm$8      & 233$\pm$7      \\
\bottomrule
\end{tabular}
\end{table}

Pitch extraction was performed for 12.9k tokens ($f_0$ could not be extracted for 1.1k tokens). We analyzed tokens in six categories: (1) fillers, (2) fillers by a semantic boundary, (3) fluent tokens, (4) tokens by a semantic boundary, (5) disfluent tokens by an interruption point, and (6) tokens within disfluencies near no semantic boundaries or interruption points. 
The $f_0$ mean and standard deviation were computed for each token; the average of these statistics were then calculated for each category (1--6) and then by speaker. We compared per-category statistics differences between gender and age groups. 
Both male and female pitch exhibited similar behaviors across most categories, with an average $f_0$ around 242Hz for fluent tokens. Female speakers showed a slight increase in mean $f_0$ for disfluent tokens, while male speakers showed a slight decrease in this category. 
An aggregate of the data as separated by age is in Table \ref{tab:age_c_averages}: mean $f_0$ for all categories decreased as age increased. 
Additionally, for both genders and across all age groups, fluent tokens (category 3) and tokens within disfluencies near no boundary points (category 6) both exhibited a lower standard deviation than the other categories. 
These values are not normalized as children have shown less regular variability in pitch than adults, likely due to physiological differences.




%% file: compare-adult.tex
\section{Comparison to Adult Speech}
\subsection{Adult Speech Corpora}
\label{ssec:adult-corpora}
We compare disfluency patterns in our child speech corpus and two adult speech corpora distributed by the Linguistics Data Consortium: Switchboard (Swbd) \cite{Godfrey1993} and CallHome \cite{Canavan1997}. Swbd is a collection of English telephone speech between two strangers who were given specific topics. A subset of Swbd has annotated disfluencies, making the corpus widely used in disfluency detection research. CallHome comprises English telephone conversations; the conversations are unscripted but most participants chose to call their family members or close friends.

\subsection{Comparative Statistics}
\label{ssec:adult-stats}
Table \ref{tab:disf-stats} summarizes disfluency statistics in the 3 datasets. The Child corpus has a significantly higher disfluency
rate and shorter average SU lengths than the adult corpora.
Including `like' as a filler, children have a higher filler rate.
It has been observed that adults use `uh' much more frequently than `um' \cite{le2017and}, but the opposite is seen for the children in our corpus. Inter-speaker rate variation is similar for children and adults.
Repair and reparandum statistics are given in Table \ref{tab:repair-reparandum-lengths}. On average, children have longer reparandums than repairs, while the opposite is true for adults. This analysis was done on simple disfluencies, excluding complex/nested disfluencies.

\begin{table}[hbpt]
\centering
\caption{Disfluency statistics across 3 datasets. {\bf Bold} denotes statistically significant difference between child speech and adult speech at $p<0.01$.}
\label{tab:disf-stats}
\begin{tabular}{llll}
\toprule
                   & Child  & CallHome & Swbd   \\
\midrule
\# tokens          & 13,568 & 43,160   & 64,944 \\
\# turns           & 2,119  & 5,869    & 8,604  \\
avg. SU length     & {\bf 6.4}	& 7.4	   & 7.5    \\
disf. rate         & {\bf 10.1\%} & 6.3\%    & 6.2\%  \\
`uh' rate          & {\bf 0.5\%}  & 0.9\%    & 2.7\%  \\
`um' rate          & {\bf 2.3\%}  & 0.6\%    & 0.5\%  \\
frag. rate         & {\bf 1.8\%}  & 1.2\%    & 0.5\%  \\
\bottomrule
\end{tabular}
\end{table}

\begin{table}[hbpt]
\centering
\caption{Average statistics of repair and reparandum lengths in 3 datasets. {\bf Bold} denotes statistically significant difference between child speech and adult speech at $p<0.01$.}
\label{tab:repair-reparandum-lengths}
\begin{tabular}{llcl}
\toprule
                             & Child & CallHome & Swbd \\
\midrule
\# of disfluent regions      & 525   & 1068     & 2159 \\
\# non-nested disfluencies   & 474   & 922      & 1923 \\
mean repair length           & {\bf 1.71}  & 2.04     & 1.90 \\
mean reparandum length       & {\bf 2.46}  & 2.11     & 1.59 \\
mean repair:reparandum ratio & {\bf 0.87}  & 1.13     & 1.25 \\
\bottomrule
\end{tabular}
\end{table} 

\subsection{Automatic Disfluency Detection}
\label{ssec:disf-detection}

For automatic disfluency detection, we use a bidirectional LSTM-CRF model with a neural pattern match network \cite{zayats2018robust}, since this model (trained on Swbd) has been shown to be robust in testing on different domains. The framework uses a BIO tagging approach, where each token is predicted to be either fluent or part of a reparandum, repair or both. Following most previous studies, the overall performance is measured in F1 score of correctly predicted disfluencies in the reparandum.

The disfluency detection results on the Child data are shown in Table \ref{tab:disfluency-detection} together with the results reported in \cite{zayats2018robust} for the adult conversation corpora. This system performs surprisingly well on the child speech, achieving an F1 score that is 10\% higher than on CallHome. 
The fact that the child speech is elicited by an unknown interviewer vs.\ a family member might explain why 
disfluency detection worked reasonably well here compared to on CallHome. 

\begin{table}[hbpt]
\caption{Disfluency detection scores across 3 datasets}
\label{tab:disfluency-detection}
\centering
\begin{tabular}{llcl}
\toprule
Measure   & Child & CallHome & Swbd \\ 
\midrule
precision & 0.85  & 0.66     & 0.93 \\
recall    & 0.70  & 0.66     & 0.83 \\
F1        & 0.77  & 0.66     & 0.88 \\
\bottomrule
\end{tabular}
\end{table}

In cases where the automatic system missed disfluencies in the Child corpus, the disfluency tends to be more complex or span over multiple tokens. Some examples are shown below, with the missed disfluent tokens underlined.
\begin{itemize}[nolistsep]
\item $[[$and to + and + and$]$ \underline{we have to clean} + $[$if + if you + if$]$ when it’s night we have to clean$]$ our teeths
\item because $[$\underline{you don’t want people to say} + when you’re talking you don’t want people to say$]$ this 
\item and you can make different colors $[$\underline{at} on- + \underline{out of} + out of$]$ two colors 
\end{itemize}


While we cannot directly compare results of different automatic detection algorithms on different corpora, 
it is notable that \cite{Yildirim09} reports roughly comparable automatic interruption point detection results using only language cues for a system trained on children's speech (F1=0.75 vs.\ F1=0.73 for our corpus).

%% file: conclusion.tex
\section{Conclusions}
We presented a novel corpus of child speech transcripts annotated with disfluencies.
Our analyses show that disfluency patterns in children are significantly different from adult speech: children have higher disfluency and filler rates, have longer reparandums than repairs, and exhibit gender differences both similar (female children have lower disfluency rates) and distinct from adults (male children have lower filler rates). 
Despite the domain mismatch, a disfluency detection system trained on adult transcripts can detect disfluencies in our corpus relatively well, achieving an F1 score of 0.77. Our acoustic analysis further shows pitch pattern differences between children by gender and age: pitch for both fluent and disfluent words reduces with age, and female pitch increases slightly from fluent to disfluent regions, while the opposite is observed for male children.
We are collecting and annotating data for a third year in this project, which will provide further data for studying age effects.
